\documentclass[a4paper, 10pt, conference]{ieeeconf}      
\usepackage{FG2026}

\FGfinalcopy 

\IEEEoverridecommandlockouts                              
\usepackage{epsfig} 
\usepackage{mathptmx} 
\usepackage{times} 
\usepackage{amsmath} 
\usepackage{amssymb}  
\usepackage{graphicx}
\usepackage{booktabs}
\let\labelindent\relax
\usepackage{enumitem}
\usepackage{verbatim}
\usepackage{colortbl}
\usepackage{pifont}
\newcommand{\xmark}{\ding{55}}   
\usepackage[table]{xcolor}
\usepackage{stfloats}
\usepackage{multirow}
\usepackage[hidelinks]{hyperref} 
\usepackage{cleveref}
\usepackage{makecell}

\def\FGPaperID{17} 

\title{\LARGE \bf
Generative Data Augmentation for Skeleton Action Recognition}

\usepackage{fancyhdr}
\author{\parbox{16cm}{\centering
    {\large Xu Dong$^1$, Wanqing Li$^2$, Anthony Adeyemi-Ejeye$^1$ and Andrew Gilbert$^1$}\\
    {\normalsize
    $^1$ Innovative Media Lab, University of Surrey, Guildford, UK\\
    $^2$ Advanced Multimedia Research Lab, University of Wollongong, Wollongong, Australia}}
}

\begin{document}

\maketitle
\thispagestyle{fancy}

\begin{abstract}

Skeleton-based human action recognition is a powerful approach for understanding human behaviour from pose data, but collecting large-scale, diverse, and well-annotated 3D skeleton datasets is both expensive and labor-intensive. To address this challenge, we propose a conditional generative pipeline for data augmentation in skeleton action recognition. Our method learns the distribution of real skeleton sequences under the constraint of action labels, enabling the synthesis of diverse and high-fidelity data. Even with limited training samples, it can effectively generate skeleton sequences and achieve competitive recognition performance in low-data scenarios, demonstrating strong generalisation in downstream tasks. Specifically, we introduce a Transformer-based encoder–decoder architecture, combined with a generative refinement module and a dropout mechanism, to balance fidelity and diversity during sampling. Experiments on HumanAct12 and the refined NTU-RGBD (NTU-VIBE) dataset show that our approach consistently improves the accuracy of multiple skeleton-based action recognition models, validating its effectiveness in both few-shot and full-data settings. The source code can be found at \href{https://github.com/dx199771/Generative-Data-Augmentation-for-Skeleton-Action-Recognition}{\textit{here}}.

\end{abstract}

\begin{figure*}[!htp]
\centering
\includegraphics[width=0.97\linewidth]{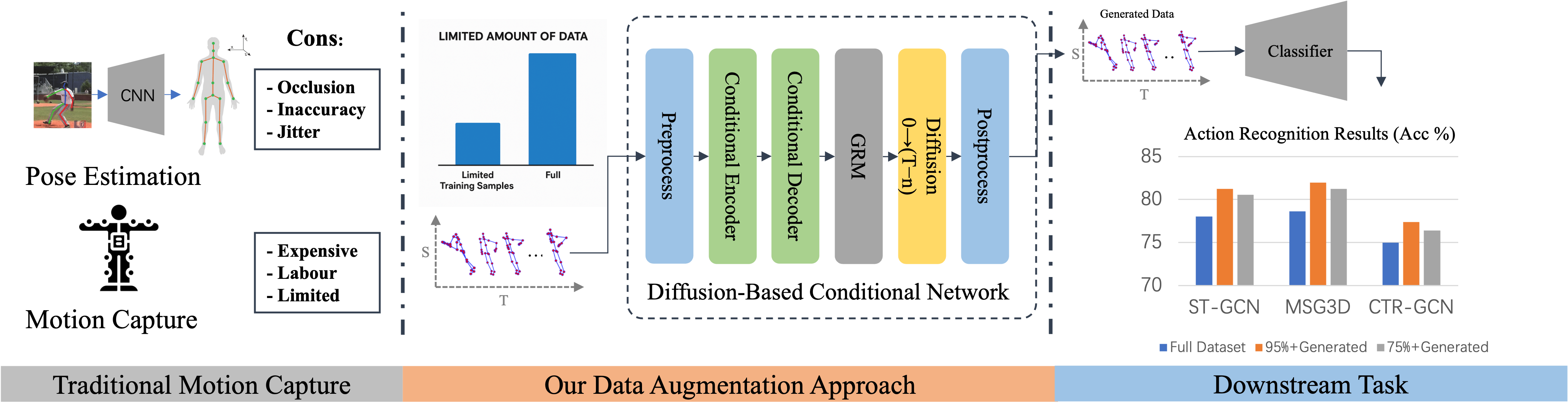}
\caption{Overview of our approach. With only a small set of labelled skeleton sequences, the model generates diverse and high-fidelity samples. When combined with a reduced amount of real data for training, these synthetic samples enable our skeleton action recognisers to achieve performance close to the state of the art on HumanAct12 and Refined NTU RGB+D.}
\label{fig:teaser}
\end{figure*}

\section{INTRODUCTION}

Human action recognition is a key task in computer vision with applications in human-computer interaction, video surveillance, healthcare, and virtual reality. Among various modalities, 3D skeleton-based action recognition has emerged as a lightweight, privacy-preserving solution.
It encodes only the positions of key joints, making it robust to appearance, lighting, and background variations, while being efficient in storage and computation.

However, acquiring large-scale, high-quality skeleton datasets remains challenging. High-precision optical motion capture systems require expensive specialised equipment, with costs often exceeding \$10,000 \cite{provini2023emerging}. Therefore, other datasets compromise by relying on depth sensors (e.g., Kinect V2) \cite{shahroudy2016nturgbdlargescale} or multi-view camera setups \cite{zou2022humanposeshapeestimation,6682899}, which still demand controlled environments and active subject participation. Moreover, even in carefully controlled settings, the captured data can still be highly cumbersome, noisy, especially from depth sensors, and resource-intensive. Deep learning based pose estimation methods can extract 3D poses from RGB inputs~\cite{kocabas2020vibevideoinferencehuman,8765346}, but the results often suffer from noise and inconsistencies, especially in unconstrained scenes.

To address the high collection cost, limited diversity, and noise in existing datasets, many approaches have explored data augmentation~\cite{electronics13040747,tu2018spatial,meng2019sample,huynh2019encoding,Cormier_2024_WACV}. These methods fall into two categories: transformation-based, which apply spatial and temporal perturbations (e.g., rotation, scaling, noise), and generation-based, which synthesise new sequences using frameworks like VAEs, GANs, or diffusion models~\cite{shen2023imaginativegenerativeadversarialnetwork,ren2023diffusionmotiongeneratetextguided}. While the former often require careful tuning of hyperparameters, the latter, while capable of learning the underlying data distribution to generate realistic samples, frequently suffers from limited diversity and a strong dependence on large-scale data. Building on MDM \cite{tevet2023human}, we introduce a conditional semantic encoder and the fidelity–diversity control module, and replace classifier-free guidance with classifier guidance during sampling to prioritise class alignment for recognition explicitly.

Specifically, as shown in \ref{fig:teaser}, this work proposes a conditional diffusion-based data augmentation method for 3D skeleton-based action recognition. Our method can efficiently generate high-fidelity, diverse, discriminative, and label-consistent skeleton data, providing both realistic variations and strong supervision signals for downstream recognition models. In the training phase, our method employs a Transformer encoder to extract the semantic information of the original skeleton data, while incorporating action labels as supervision signals. The Transformer decoder takes the noise tokens together with the conditional representation, which integrates semantic features, temporal information, and action labels. Guided by both reconstruction and classification objectives, it progressively denoises the tokens into label-consistent skeleton sequences, while jointly capturing structural priors and maintaining label consistency. In the inference phase, the diffusion model generates realistic and label-consistent skeleton sequences. To further improve generation quality, we design a Generative Refinement Module (GRM) and introduce a sampling-time dropout mechanism to balance fidelity and diversity, encouraging the model to produce discriminative and label-consistent variations. Our method is highly efficient, requiring only a single training phase. Once trained, the model is capable of generating large-scale skeleton data during inference, while allowing explicit control over the trade-off between diversity and fidelity in the generated samples.

We conduct extensive experiments on HumanAct12~\cite{guo2020action2motion} and Refined NTU-RGBD (NTU-VIBE) \cite{shahroudy2016nturgbdlargescale,guo2020action2motion}, evaluating generation quality and downstream recognition performance. Our method demonstrates strong generalisation, particularly in low-data scenarios, where adding synthetic samples significantly improves accuracy, reaching levels comparable to those achieved with full data training. We conducted comprehensive experiments to evaluate the effectiveness of our method. By assessing the generation results and the performance on downstream skeleton-based action recognition tasks, we demonstrated the superior performance of our approach. In scenarios with limited real data, adding synthetic samples significantly improves accuracy, reaching levels comparable to using more real data that would be expensive and challenging to obtain. Furthermore, we conducted experiments to optimise the augmentation process by balancing diversity and fidelity in the synthetic data. Moreover, our data augmentation method demonstrates strong generality, as it can be adapted to various skeleton data formats and is compatible with a wide range of skeleton-based action recognition datasets and methods.

\noindent \textbf{Contributions:}
\begin{itemize}
\item We propose a conditional skeleton generation method based on diffusion models, conditioned on action labels, to generate diverse and realistic motion sequences. By generating large amounts of high-quality data from limited training samples, our approach reduces the need for costly large-scale data collection.
\item We introduce a transformer encoder that extracts semantic representations from skeleton inputs and incorporates action labels as conditional signals to guide the diffusion-based generation process.
\item We introduce a Generative Refinement Module (GRM) and sampling-time dropout to control fidelity and diversity in the synthetic data jointly.
\item We validate our method across two datasets and multiple skeleton action recognition backbones, showing improvements in both few-shot and full-data training scenarios. Additionally, we conduct ablation studies to evaluate the contribution of each module and assess the quality of generated skeletons using standard metrics.
\end{itemize}

\section{Related Work}
\label{sec:relatework}

\subsection{Diffusion Models.}
Diffusion models \cite{sohldickstein2015deepunsupervisedlearningusing,song2022denoisingdiffusionimplicitmodels} are generative models that produce data by learning to reverse a progressive noising process. Denoising Diffusion Probabilistic Models (DDPM) \cite{ho2020denoisingdiffusionprobabilisticmodels,song2022denoisingdiffusionimplicitmodels} and Denoising Diffusion Implicit Models (DDIM) have demonstrated state-of-the-art results in image generation. Conditional diffusion techniques, such as classifier guidance \cite{dhariwal2021diffusionmodelsbeatgans} and classifier-free guidance \cite{ho2022classifierfreediffusionguidance}, enable fine-grained control during sampling. Beyond images, diffusion models have shown strong potential in motion generation tasks. Human motion is typically represented as sequences of joint data in 2D, 3D, or SMPL \cite{SMPL:2015,zhu2023humanmotiongenerationsurvey}. Recent works \cite{tevet2023human,chen2023executing,ren2023realistic,dabral2023mofusionframeworkdenoisingdiffusionbasedmotion,kim2023flamefreeformlanguagebasedmotion,liu2024dgfm} have shown strong success in synthesising realistic, diverse, and controllable motion sequences. While diffusion models have been explored for motion generation, no prior work has applied conditional diffusion for label-guided skeleton augmentation in recognition pipelines.

\subsection{Synthetic Data for Augmentation.}
Data scarcity often leads to overfitting and poor generalisation in neural networks, especially under low-data regimes. Traditional augmentation methods \cite{krizhevsky2012imagenet} introduce simple transformations (e.g., flips, noise, crops) but are limited in diversity. Generative approaches overcome this by learning data distributions to produce new samples. Early work like DAGAN \cite{antoniou2018dataaugmentationgenerativeadversarial} and BigGAN \cite{brock2018large} explored this idea to generate diverse image data for improving classification tasks. More recent efforts leverage text-to-image diffusion models. The study by Jahanian et al.~\cite{jahanian2022generativemodelsdatasource} explored the feasibility of learning general-purpose visual representations from generative models instead of relying solely on original data. With the rapid development of diffusion models in recent years, this technology has become a new trend in generating training data, benefiting from its stationary training objective, high diversity, and conditional generation capabilities. \cite{trabucco2023effectivedataaugmentationdiffusion} proposed DA-Fusion that utilised a large pre-trained text-to-image diffusion model to address the weaknesses of standard data augmentation while retaining the strengths. For skeleton data, augmentation is less explored. \cite{10689680} analyses synthetic data on the fall-down detection task. \cite{Cormier_2024_WACV} proposed a skeleton data augmentation method derived from observations of inaccuracies in human pose estimation. The works apply geometric perturbations (e.g., rotation, translation) or simulate occlusion. However, most do not model the complex distribution of temporal joint sequences. To our knowledge, this is the first work to apply conditional diffusion models for class-aware skeleton data augmentation, enabling label consistent generation at scale.

\subsection{Skeleton Action Recognition.}
Early skeleton recognition methods relied on handcrafted features and classical classifiers \cite{hu2015jointly,wu2014leveraging}, but they require manual feature design, are sensitive to noise/viewpoint changes, and poorly capture long-range dynamics. With the development of deep learning, recognition has shifted from handcrafted pipelines to end-to-end RNN/GCN/Transformer architectures that learn robust spatiotemporal representations from raw skeletons. Graph Convolutional Networks (GCNs) became the standard due to the ability to model the spatial and temporal relationships of skeleton data effectively. ST-GCN \cite{yan2018spatialtemporalgraphconvolutional} introduced spatial-temporal graphs but incurred a high computational cost. MSG3D \cite{liu2020disentangling} captured multi-scale patterns; CTR-GCN \cite{chen2021channelwisetopologyrefinementgraph} used channel-wise topology refinement to learn adaptive topologies and aggregates joint features for dynamic structure learning; BlockGCN \cite{zhou2024blockgcn} simplified the graph via blockwise partitioning,  performing independent modelling within each block but with limited temporal modelling. We use these models as baselines to evaluate the benefit of our synthetic data.

\begin{figure*}[!htbp]
\centering
\includegraphics[width=0.99\linewidth]{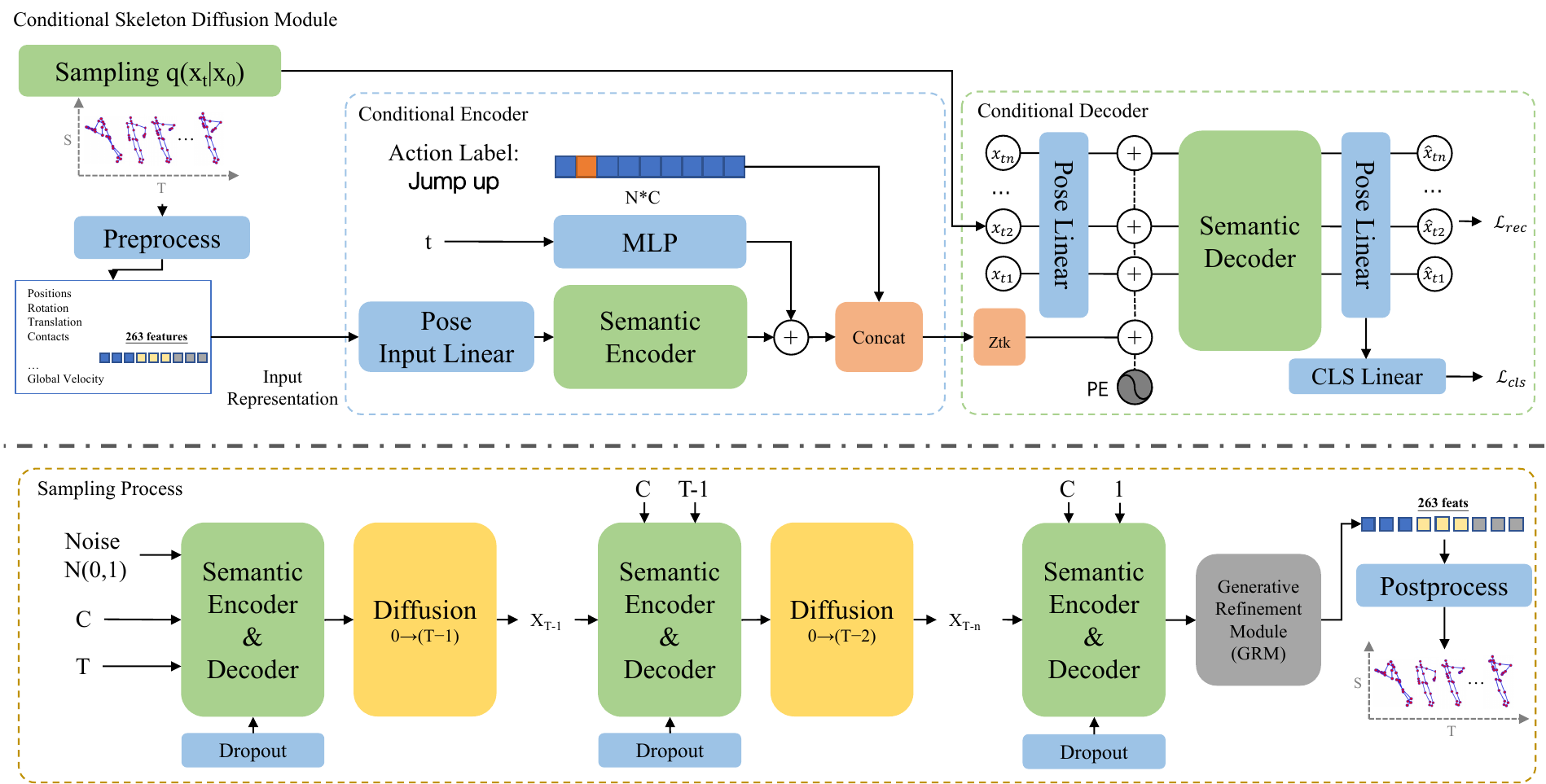}
\caption{Overview of our proposed network. \textbf{(Top) Conditional Skeleton Diffusion Module.} The encoder processes a skeleton feature sequence together with the noise step $t$ and the corresponding action label, producing a conditional representation. A Transformer-based decoder then reconstructs the clean skeleton sequence from the noise-corrupted input, guided by this representation. In addition, a lightweight classification network is introduced to encourage label-consistent generation. \textbf{(Bottom) Sampling Process.} The sampling input consists of the conditional representation and random noise, where the noise incorporates both label information and semantics from the original data. The decoder progressively denoises the sequence from step $T$ to $1$, generating a clean skeleton motion. A Generative Refinement Module and Dropout further enhance the balance between semantic fidelity to the action and diversity of the generated motions.}
\label{fig:network}
\end{figure*}

\section{Methodology}
\subsection{Diffusion Models Preliminary}
Diffusion models~\cite{ho2020denoisingdiffusionprobabilisticmodels,nichol2021improveddenoisingdiffusionprobabilistic,rombach2022highresolutionimagesynthesislatent} are generative frameworks that learn data distributions by simulating a forward process that gradually adds Gaussian noise, and a reverse process that removes it.
In the forward process, the posterior distribution is implemented as a Markov chain that recursively adds noise to the sample through the conditional probability. This process can be denoted as:
 
\noindent
\begin{equation}
q(\mathbf{x}_{1:T} | \mathbf{x}_0) = \prod_{t=1}^T q(\mathbf{x}_t | \mathbf{x}_{t-1}),
\end{equation}
\hfill
\begin{equation}
q(\mathbf{x}_t | \mathbf{x}_{t-1}) = \mathcal{N}(\mathbf{x}_t; \sqrt{1-\beta_t} \mathbf{x}_{t-1}, \beta_t \mathbf{I}),
\end{equation}
 
Here, $\beta_t$ is a variance schedule, $\alpha_t = 1 - \beta_t$ represents the proportion of signal retained at step $t$ and $\mathbf{I}$ denotes the identity matrix of appropriate dimensions. Given a timestep $t$, $\mathbf{x}_t$ can be sampled directly from $\mathbf{x}_0$ as:
\begin{equation}
\mathbf{x}_t = \sqrt{\bar{\alpha}_t} \mathbf{x}_0 + \sqrt{1 - \bar{\alpha}_t} \boldsymbol{\epsilon},
\end{equation}

where $\boldsymbol{\epsilon} \sim \mathcal{N}(\mathbf{0}, \mathbf{I})$ is a standard Gaussian noise term, and $\bar{\alpha}_t = \prod_{s=1}^t \alpha_s$ denotes the cumulative signal strength. Rather than predicting the noise $\boldsymbol{\epsilon}$, we follow recent paradigms~\cite{tevet2023human,ramesh2022hierarchicaltextconditionalimagegeneration} to directly predict the original sample $\mathbf{x}_0$ from the noisy input. The training objective is formulated as:


\begin{equation}
L = \mathbb{E}_{\mathbf{x}_0 \sim q(\mathbf{x}_0|\mathbf{c}), t \sim [1, T]} 
\left[ \|\mathbf{x}_0 - G(\mathbf{x}_t, t, \mathbf{c})\|_2^2 \right],
\end{equation}

where $t$ and $\mathbf{c}$ denote the timestamps and condition information, respectively.

\subsection{Conditional Diffusion Model}
An overview of our pipeline is illustrated in \Cref{fig:network}. The architecture is a Transformer model with action labels as conditioning signals. The model includes
\begin{itemize}
    \item  \textbf{Conditional Encoder:} A Transformer encoder to extract latent feature representations from the skeleton input data along with \textbf{timestep and label embeddings.}
    
    \item \textbf{Conditional Decoder:} A Transformer decoder reconstructs the original skeleton data, taking the encoder features, concatenated with the skeleton data corrupted by noise through the diffusion process.
\end{itemize}
During sampling, we use a Generative Refinement Module (GRM) to discard low-fidelity generations and apply dropout to promote diversity further, ensuring that the final output is both discriminative and robust for downstream tasks.

\subsection{Input Representation.}
Skeleton data is compact but semantically rich. In our setting, the original HumanAct12 dataset provides 3D coordinates for 22 skeletal joints. Following the HumanML3D representation~\cite{Guo_2022_CVPR}, we convert each frame into a 263-dimensional feature vector, where the 22 joints are re-encoded to jointly capture 3D positions, local orientations, and dynamic attributes such as velocities. This extended representation offers a more comprehensive description of human motion, preserving both spatial configurations and temporal dynamics, while remaining computationally efficient compared to raw mesh or video data. Detailed construction of the 263-dimensional features is provided in \ref{tab:feature_vector}.

\begin{table}[t]
\centering
\caption{The 263-dimensional feature vector explanations.}
\begin{tabular}{lc p{3.5cm}}
\toprule
\textbf{Component} & \textbf{Dimensions} & \textbf{Description} \\
\midrule
Joint Positions   & $22 \times 3 = 66$  & 3D coordinates $(x, y, z)$ for 22 joints \\
\midrule
Joint Velocities  & $22 \times 3 = 66$  & Velocity vectors for each joint \\
\midrule
Joint Rotations   & $22 \times 6 = 132$ & 6D rotation representations (more stable than 
quaternions or Euler angles) \\\midrule
Global Translation& $3$                 & Overall body translation in 3D space \\\midrule
Global Velocity   & $3$                 & Global movement velocity of the body \\\midrule
\midrule
\textbf{Total}    & \textbf{263}        & Combined total of all feature components \\
\bottomrule
\end{tabular}
\label{tab:feature_vector}
\end{table}

\subsection{Conditional Encoder}
The input 263-dimensional feature sequence is first processed with temporal positional embeddings to preserve frame-wise order information. The action label $c$ is represented as a one-hot vector and embedded through an MLP, while the diffusion timestep $t$ is similarly mapped into the latent space. These conditional embeddings are concatenated and projected as a prefix token $z_{tk}$, which is then prepended to the feature sequence and fed into the encoder. The conditional encoder allows the model to incorporate both semantic (action label) and temporal (timestep) guidance during representation learning.

\subsection{Conditional Decoder}

The decoder takes the noisy feature sequence together with the conditional prefix token $z_{tk}$ and performs token-level self-attention to reconstruct the underlying motion dynamics. It outputs a denoised 263-dimensional feature sequence, which is then passed through a 2-layer MLP classifier to predict the action label. This auxiliary classification objective provides label supervision, ensuring that the generated motion not only reduces diffusion noise but also remains consistent with the intended action semantics.

\subsection{Sampling Process}
Our sampling involves predicting the clean sample $\hat{x}_0$ at each time step $t$, and then adding noise to regress it back to $x_{t-1}$. This iterative process continues from $t = T$ until $t = 0$, producing the final sample $x_0$. Unlike previous work \cite{shafir2023humanmotiondiffusiongenerative, tevet2023human}, which uses classifier-free guidance (occasionally masking conditions), we condition explicitly on labels throughout training and sampling, as fidelity to specific actions is essential for data augmentation. To encourage sample diversity and prevent the model from overfitting to label-conditioned patterns, we apply dropout within the denoising network during the sampling process. The stochasticity introduced in token activations allows our model to take a single action label as input and generate multiple diverse motion sequences with subtle variations not only in joint dynamics but also in higher-level semantics such as speed, thereby enriching data diversity without the need for extensive skeleton data collection.

\subsection{Generative Refinement Module (GRM)}
The GRM evaluates generated samples $\hat{x}_0$ using a deviation measure $d(\hat{x}_0, x_0)$. 
Samples exceeding the threshold $\tau$ are discarded, and the retained set is defined as
\begin{equation}
\mathcal{S} = \{\, \hat{x}_0 \;\mid\; d(\hat{x}_0, x_0) \leq \tau \,\},
\label{eq:grm}
\end{equation}
where $x_0$ denotes the reference ground-truth sample (or its conditional embedding), 
$d(\cdot)$ is the deviation metric (e.g., $\ell_2$ distance in the 263-dimensional feature space), 
and $\tau$ is the deviation threshold. 
This filtering ensures that retained samples remain close to the real distribution (fidelity), 
while the combination with sampling-time dropout introduces diverse yet label-consistent variations.

\subsection{Loss function}
Our total loss combines a \textbf{Reconstruction loss} and  \textbf{Classification loss}. The Reconstruction loss $\mathcal{L}_{rec}$ enforces the generated samples to match the target data in the integrated 263-dimensional feature space. where $G(x_t, t, c)$ is the generated skeleton and $x_0$ is the ground truth. The Classification loss $\mathcal{L}_{cls}$ is a cross-entropy loss applied to the predicted action class of the generated data.
\begin{equation}
\mathcal{L}_{\text{rec}} = \mathbb{E}_{x_0, t} \left[ \| x_0 - G(x_t, t, c) \|_2^2 \right]
\end{equation}

\begin{equation}
\mathcal{L}_{\text{cls}} = - \frac{1}{N} \sum_{i=1}^{N} \log \sigma_{y_i}(\mathbf{f}_i)
\end{equation}
 
Where $\mathbf{f}i$ denotes the predicted logits for the $i$-th sample, and $\sigma{y_i}(\mathbf{f}_i)$ represents the predicted probability for the ground-truth class label $y_i$, obtained via the softmax function applied to $\mathbf{f}_i$. The total loss $\mathcal{L}$ adopts a weighted combination of the reconstruction loss and the classification loss, where $\lambda$ is a weighting hyperparameter used to balance.

\begin{equation}
\mathcal{L} = \mathcal{L}_{\text{rec}} + \lambda_{\text{cls}} \mathcal{L}_{\text{cls}}
\end{equation}

\begin{table*}[htbp]
\centering
\caption{Comparison on \textbf{HumanAct12} using skeleton-based action recognition models. Results are reported as \emph{mean} ± \emph{std} over 5 independent runs; Methods marked with * denote models trained on augmented data (real + synthetic). Improvements brought by our augmented data are highlighted in \textcolor{green!50!black}{green}.}
\label{tab:augresult}

\DeclareRobustCommand{\std}[1]{\,{\begingroup\color{black!65}\scriptsize$\pm$#1\endgroup}}

\normalsize
\begin{tabular}{lcccc}
\toprule
\multirow{2}{*}{Method} & \multicolumn{4}{c}{Real Data Usage} \\
\cmidrule(lr){2-5}
 & 100\% & 95\% & 90\% & 75\% \\
\midrule

STGCN++ \cite{duan2022pyskl} 
& \cellcolor{gray!20}78.47\std{2.09}
& \cellcolor{gray!20}77.78\std{2.55}
& \cellcolor{gray!20}75.83\std{1.24}
& \cellcolor{gray!20}73.89\std{0.38} \\
STGCN++* 
& \cellcolor{green!20}83.19\std{2.73} \textcolor{gray}{\scriptsize (↑4.72)} 
& \cellcolor{green!20}81.63\std{2.05} \textcolor{gray}{\scriptsize (↑3.85)} 
& \cellcolor{green!20}81.50\std{1.47} \textcolor{gray}{\scriptsize (↑5.66)} 
& \cellcolor{green!20}81.11\std{0.80} \textcolor{gray}{\scriptsize (↑7.22)} \\
\midrule
MSG3D \cite{liu2020disentangling} 
& \cellcolor{gray!20}80.42\std{1.99}
& \cellcolor{gray!20}77.64\std{1.50}
& \cellcolor{gray!20}76.94\std{2.43}
& \cellcolor{gray!20}74.86\std{1.80} \\
MSG3D* 
& \cellcolor{green!20}83.11\std{3.46} \textcolor{gray}{\scriptsize (↑2.69)} 
& \cellcolor{green!20}83.24\std{1.23} \textcolor{gray}{\scriptsize (↑5.60)} 
& \cellcolor{green!20}81.77\std{1.18} \textcolor{gray}{\scriptsize (↑4.83)} 
& \cellcolor{green!20}80.50\std{0.68} \textcolor{gray}{\scriptsize (↑5.64)} \\
\midrule
CTRGCN \cite{chen2021channelwisetopologyrefinementgraph}
& \cellcolor{gray!20}77.78\std{1.97}
& \cellcolor{gray!20}76.94\std{2.10}
& \cellcolor{gray!20}75.56\std{1.42}
& \cellcolor{gray!20}73.61\std{2.41} \\
CTRGCN* 
& \cellcolor{green!20}79.42\std{2.02} \textcolor{gray}{\scriptsize (↑1.64)} 
& \cellcolor{green!20}79.59\std{1.83} \textcolor{gray}{\scriptsize (↑2.65)} 
& \cellcolor{green!20}80.16\std{2.20} \textcolor{gray}{\scriptsize (↑4.60)} 
& \cellcolor{green!20}78.25\std{1.72} \textcolor{gray}{\scriptsize (↑4.64)} \\
\midrule
BlockGCN \cite{zhou2024blockgcn}
& \cellcolor{gray!20}77.78\std{1.30}
& \cellcolor{gray!20}75.67\std{1.30}
& \cellcolor{gray!20}75.56\std{0.76}
& \cellcolor{gray!20}75.56\std{0.90} \\
BlockGCN* 
& \cellcolor{green!20}78.91\std{0.41} \textcolor{gray}{\scriptsize (↑1.13)} 
& \cellcolor{green!20}78.67\std{1.63} \textcolor{gray}{\scriptsize (↑3.00)} 
& \cellcolor{green!20}78.19\std{0.38} \textcolor{gray}{\scriptsize (↑2.63)} 
& \cellcolor{green!20}77.17\std{0.72} \textcolor{gray}{\scriptsize (↑1.61)} \\
\bottomrule
\end{tabular}
\end{table*}

\begin{table*}[htbp]
\centering

\caption{Comparison on the \textbf{Refined NTU RGB+D} dataset using skeleton-based action recognition models. Results are reported as \emph{mean} ± \emph{std} over 5 independent runs; Methods marked with * denote models trained on augmented data (real + synthetic). Improvements from our augmented data are highlighted in \textcolor{green!50!black}{green}.}
\label{tab:augresult2}
\DeclareRobustCommand{\std}[1]{\,{\begingroup\color{black!65}\scriptsize$\pm$#1\endgroup}}

\normalsize
\begin{tabular}{lcccc}
\toprule
\multirow{2}{*}{Method} & \multicolumn{4}{c}{Real Data Usage} \\
\cmidrule(lr){2-5}

 & 25\% & 20\% & 15\% & 10\% \\
\midrule

STGCN++ \cite{duan2022pyskl}
& \cellcolor{gray!20}91.55\std{0.62}
& \cellcolor{gray!20}90.95\std{1.04}
& \cellcolor{gray!20}89.94\std{1.06}
& \cellcolor{gray!20}83.01\std{2.15} \\
STGCN++$^{*}$
& \cellcolor{green!20}92.36\std{0.33} \textcolor{gray}{\scriptsize (↑0.81)} 
& \cellcolor{green!20}92.14\std{0.87} \textcolor{gray}{\scriptsize (↑1.18)} 
& \cellcolor{green!20}92.07\std{0.76} \textcolor{gray}{\scriptsize (↑2.13)} 
& \cellcolor{green!20}85.38\std{1.13} \textcolor{gray}{\scriptsize (↑2.37)} \\
\midrule
MSG3D \cite{liu2020disentangling}
& \cellcolor{gray!20}90.97\std{1.08}
& \cellcolor{gray!20}89.74\std{2.33}
& \cellcolor{gray!20}87.41\std{1.30}
& \cellcolor{gray!20}79.48\std{1.87} \\
MSG3D$^{*}$
& \cellcolor{green!20}92.30\std{0.39} \textcolor{gray}{\scriptsize (↑1.33)} 
& \cellcolor{green!20}90.36\std{0.68} \textcolor{gray}{\scriptsize (↑0.62)} 
& \cellcolor{green!20}89.90\std{1.59} \textcolor{gray}{\scriptsize (↑2.49)} 
& \cellcolor{green!20}83.17\std{1.13} \textcolor{gray}{\scriptsize (↑3.69)} \\
\midrule
CTRGCN \cite{chen2021channelwisetopologyrefinementgraph}
& \cellcolor{gray!20}90.81\std{1.07}
& \cellcolor{gray!20}90.78\std{0.20}
& \cellcolor{gray!20}87.57\std{2.69}
& \cellcolor{gray!20}79.28\std{1.46} \\
CTRGCN$^{*}$
& \cellcolor{green!20}91.13\std{1.34} \textcolor{gray}{\scriptsize (↑0.32)} 
& \cellcolor{green!20}90.97\std{0.49} \textcolor{gray}{\scriptsize (↑0.19)} 
& \cellcolor{green!20}89.45\std{0.35} \textcolor{gray}{\scriptsize (↑1.88)} 
& \cellcolor{green!20}83.17\std{1.34} \textcolor{gray}{\scriptsize (↑3.89)} \\

\midrule
BlockGCN \cite{zhou2024blockgcn}
& \cellcolor{gray!20}90.03\std{0.72}
& \cellcolor{gray!20}88.51\std{1.11}
& \cellcolor{gray!20}86.70\std{1.46}
& \cellcolor{gray!20}75.05\std{1.43} \\
BlockGCN* 
& \cellcolor{green!20}90.91\std{0.54} \textcolor{gray}{\scriptsize (↑0.88)} 
& \cellcolor{green!20}89.13\std{1.16} \textcolor{gray}{\scriptsize (↑0.62)} 
& \cellcolor{red!20}86.05\std{1.42} \textcolor{gray}{\scriptsize (↓0.65)} 
& \cellcolor{green!20}84.43\std{0.72} \textcolor{gray}{\scriptsize (↑9.38)} \\
\bottomrule
\end{tabular}
\end{table*}

\section{Experiments and Results}
\subsection{Datasets.}
We evaluated our method on two benchmark datasets: HumanAct12~\cite{guo2020action2motion} and the Refined NTU-RGBD (NTU-VIBE)~\cite{shahroudy2016nturgbdlargescale,guo2020action2motion}.

\begin{itemize}
\item \textbf{HumanAct12} was a high-quality motion dataset derived from PHSPD~\cite{zou20203dhumanshapereconstruction,zou2020polarizationhumanshapepose}. It contained 1,191 motion clips and over 90,000 frames across 34 fine-grained action categories. Actions included detailed labels such as \textit{lift dumbbell with right hand} and \textit{drink bottle left hand}, enabling conditional generation with strong label guidance.

\item \textbf{Refined NTU-RGBD} was an improved version of the original NTU-RGBD~\cite{shahroudy2016nturgbdlargescale} dataset, with 3D joint annotations recomputed using the VIBE~\cite{kocabas2020vibevideoinferencehuman} method for better consistency and realism. It included 3,902 motion clips across 13 action categories, such as \textit{squat down}, \textit{sitting down}, and \textit{throw}.
\end{itemize}

Although our experiments focus on HumanAct12 and NTU-VIBE, our method is architecture- and dataset-agnostic, and could be extended to other skeleton-based datasets with minimal modification.

\vspace{1em}
\subsection{Usage Protocols.}
For HumanAct12, we conducted experiments under \textbf{different data availability settings} by randomly sampling 75\%, 90\%, 95\%, and 100\% of the original training data to train the diffusion model. For downstream evaluation, we then augmented the selected real data with 5× synthetic samples generated by the diffusion model and tested the recognition accuracy on the validation set.

For Refined NTU-RGBD, we also considered \textbf{different data availability settings} in the few-shot regime, using only 10\%, 15\%, 20\%, and 25\% of the original training data to train the diffusion model. Each subset was similarly augmented with 5× synthetic samples generated by our method during downstream evaluation. This demonstrates that even with limited real data, supplementing with label-consistent synthetic sequences can significantly improve recognition accuracy and approach the performance achieved with substantially larger real datasets.

We conclude that using 5× synthetic data represents a practical trade-off: generating substantially more data leads to redundancy and slows down the generation process, while too few synthetic samples provide only limited performance gains.
\vspace{1em}

\subsection{Implementation Details.}
Each skeleton sequence was loaded and filtered through preprocessing to ensure a minimum length. Motion data was normalised using the dataset-specific mean and standard deviation, and randomly cropped to a fixed-length window of $T = 48$ frames during training. Action labels were converted to 13-way or 34-way one-hot vectors based on predefined NTU and Humanact12 classes.
We trained our method with Adam using a learning rate of $1 \times 10^{-4}$, which was decreased by 0.1 at each step. Training was conducted for 600 epochs on a single NVIDIA RTX 3090 with a batch size of 256.
The Transformer encoder and decoder each consisted of 4 layers and 4 attention heads. For downstream action recognition, we used STGCN++\cite{duan2022pyskl}, MSG3D\cite{liu2020disentangling}, CTRGCN~\cite{chen2021channelwisetopologyrefinementgraph}, and BlockGCN~\cite{zhou2024blockgcn}, applying their default configurations.

\subsection{Data Augmentation Evaluation}
We evaluated performance by comparing the classification accuracy of the four state-of-the-art skeleton action recognition models, trained on real data only \textit{and} reduced amounts of real data supplemented with our synthetic data. The results showed consistent gains, particularly in low-data settings. \Cref{tab:augresult,tab:augresult2} (HumanAct12 and NTU-VIBE, respectively) highlight accuracy gains achieved by augmenting the real data with our generated data. Performance is most pronounced when less real data is used, validating the value of our approach in few-shot contexts. On the HumanAct12 dataset, we observed that while using 95\% or 90\% of real data yields reasonable performance, training with only 75\% real data augmented by our synthetic samples surpasses them, and even outperforms the model trained on 100\% real data, demonstrating superior data efficiency and augmentation quality. On the NTU-VIBE dataset, BlockGCN with 15\% data performs slightly worse,  which may be due to overfitting from synthetic data. We observe consistent improvements across backbones with varying capacity — from lightweight STGCN++ to deeper BlockGCN — highlighting the generality of our approach.

\begin{figure}[h]
    \centering
    \includegraphics[width=0.8\linewidth]{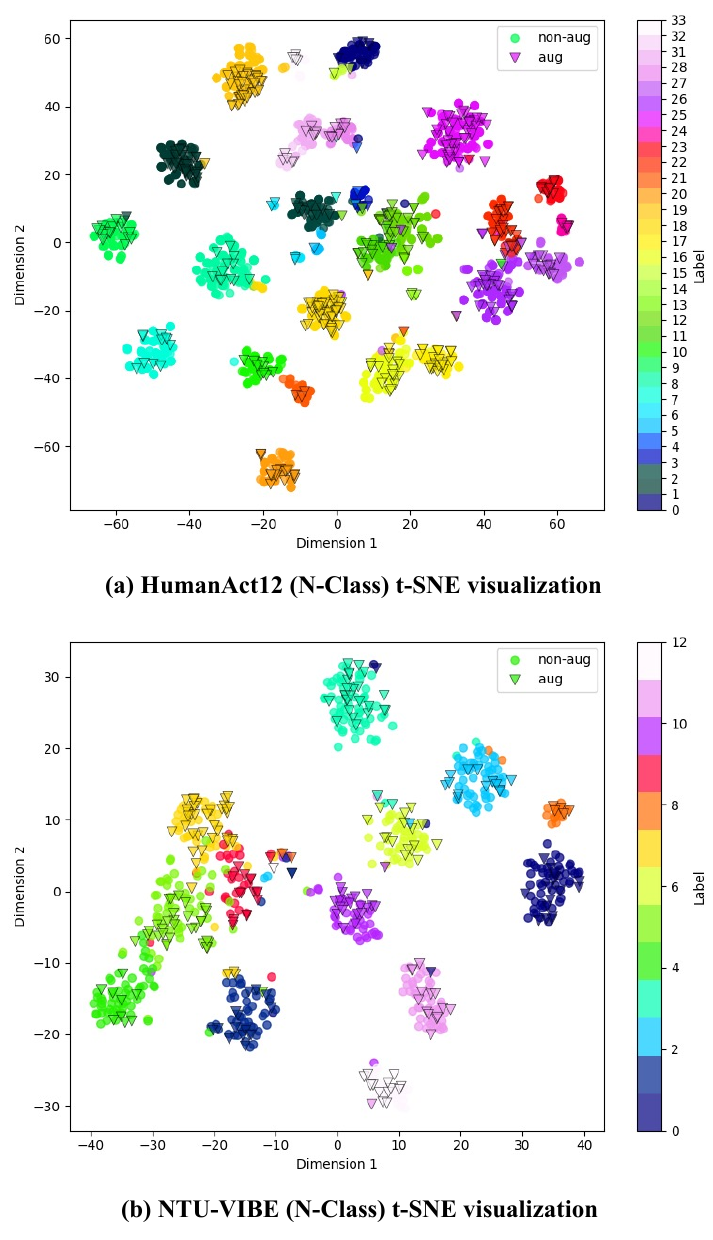}
    \caption{t-SNE visualisations comparing real and synthetic skeleton samples of (a) HumanAct12 and (b) Refined NTU-RGBD (NTU-VIBE). Real samples are denoted by ($\circ$), and synthetic samples are denoted by ($\triangledown$). }

    \label{fig:tsne}
\end{figure}
\subsection{Data Distribution Evaluation}

To examine whether our conditional diffusion augments the dataset in a label-consistent manner, we compare the distribution of \emph{real} vs. \emph{synthetic} samples using a t-SNE 2D projection. As shown in Figure~\ref{fig:tsne}, real samples are shown as circles ($\circ$), while synthetic samples are shown as inverted triangles ($\triangledown$). We observe that the synthetic samples densely populate the existing class regions, effectively enlarging per-class coverage without shifting the class centroids, thereby preserving fidelity. At the same time, some synthetic points appear along the edges of the clusters, filling low-density areas that are under-represented in the real set; this expands intra-class boundaries while remaining label-consistent, yielding greater diversity and improved generalisation in few-shot regimes. To quantify this effect, we compute the \emph{average within-class cluster covariance}, i.e., the trace of the per-class covariance matrix averaged across classes, which reflects intra-class dispersion. As shown in Table~\ref{tab:cluster_cov}, covariance increases on both datasets when synthetic samples are added, indicating broader intra-class coverage while preserving label consistency. Overall, the visualisation and quantitative results indicate that our generator adds both quantity and diversity while maintaining clear inter-class separation.

\begin{table}[h!]
\centering
\caption{Comparison of average within-class cluster covariance before and after applying data augmentation. }
\label{tab:cluster_cov}
\small
\begin{tabular}{lccc}
\toprule
\textbf{Dataset} & \textbf{Original} & \textbf{+Aug} & \textbf{Difference} \\
\midrule
HumanAct12  & 10.926 & 11.490 & +0.564 \\
NTU-VIBE    & 4.392  & 4.753  & +0.361 \\
\bottomrule
\end{tabular}
\end{table}

\subsection{Reconstruction Evaluation}
To evaluate skeleton generation quality, we compared our method with MDM~\cite{tevet2023human} and T2M-GPT~\cite{zhang2023t2mgptgeneratinghumanmotion}, using four metrics: FID (Fréchet Inception Distance), KID (Kernel Inception Distance), Diversity, and Precision/Recall. Detailed explanations of these metrics were provided in the supplementary materials. As shown in \Cref{tab:generative}, our method achieves the lowest FID and comparable KID, indicating that the generated motions are most similar to real data regarding overall distribution and visual coherence. The highest diversity among generative models demonstrates a strong ability to produce a wide range of motion styles rather than repetitive patterns. While T2M-GPT slightly outperforms Precision, suggesting high realism in individual samples, our method maintains a strong balance across all metrics. These results indicate that our approach generates realistic motions and captures a broader spectrum of plausible human movements, outperforming prior methods in fidelity and diversity.



\subsection{Ablation Study }

\subsubsection{Evaluation of proposed modules}

\begin{table}[!tbp]
\centering
\caption{Ablation study of STGCN++ results on the HumanAct12 dataset using the validation set, trained with 100\% of the data, under different combinations of modules. Results are reported as \emph{mean} ± \emph{std} over 5 independent runs.}
\DeclareRobustCommand{\std}[1]{\,{\begingroup\color{black!65}\scriptsize$\pm$#1\endgroup}}

\label{tab:augment}
\resizebox{\linewidth}{!}{   
\footnotesize
\begin{tabular}{l|cccc|c}
\toprule
{Module} & {Condition} & {CLS Loss} & {Dropout} & {Refinement} & {STGCN++ Acc.} \\
\midrule
Baseline          & \xmark & \xmark & \xmark & \xmark & 78.77\std{2.65} \\
             & \checkmark & \xmark & \xmark & \xmark & 80.62\std{1.58} \\
        & \checkmark & \checkmark & \xmark & \xmark & 81.98\std{2.56} \\
  & \checkmark & \checkmark & \checkmark & \xmark & 80.77\std{2.67} \\ \midrule
All (Ours)        & \checkmark & \checkmark & \checkmark & \checkmark & \cellcolor{green!20} \textbf{83.19}\std{2.73} \\
\bottomrule
\end{tabular}
}
\end{table}

\begin{table}[bp]
\centering
\caption{Skeleton-based action recognition performance on HumanAct12 under different Dropout values. Green cells indicate the best performance per column.}
\label{tab:dropout}
\normalsize
\renewcommand{\arraystretch}{1.0}
\vspace{0.5em}
\begin{tabular}{c|cccc}
\toprule
\textbf{Ratio} & \textbf{100\%} & \textbf{95\%} & \textbf{90\%} & \textbf{75\%} \\
\midrule
Dropout 0   & 83.85 & 84.62 & 80.77 & 83.85 \\
Dropout 0.1 & 82.31 & 82.31 & \cellcolor{green!20}\textbf{83.85} & 80.00 \\
Dropout 0.2 & \cellcolor{green!20}\textbf{86.15} & \cellcolor{green!20}\textbf{86.15} & 80.00 & \cellcolor{green!20}\textbf{84.62} \\
Dropout 0.5 & 84.62 & 84.62 & 83.08 & 81.54 \\
\bottomrule
\end{tabular}
\end{table}

\begin{table}[bp]
\centering
\caption{Skeleton-based action recognition performance on HumanAct12 under different GRM re-noise values. Green cells indicate the best performance per column.}
\label{tab:grm}
\normalsize
\renewcommand{\arraystretch}{1.0}
\vspace{0.5em}
\begin{tabular}{c|cccc}
\toprule
\textbf{Ratio} & \textbf{100\%} & \textbf{95\%} & \textbf{90\%} & \textbf{75\%} \\
\midrule
Renoise 1  & 80.77 & 83.08 & 81.54 & 85.38 \\
Renoise 2  & 83.08 & 83.08 & \cellcolor{green!20}\textbf{84.62} & 82.31 \\
Renoise 3  & 83.08 & 81.54 & 82.31 & 81.54 \\
Renoise 5  & 82.31 & 81.54 & 83.08 & 78.46 \\
Renoise 10 & 83.85 & \cellcolor{green!20}\textbf{84.62} & 80.77 & 80.00 \\
Renoise 20 & \cellcolor{green!20}\textbf{85.38} & 83.85 & 80.77 & \cellcolor{green!20}\textbf{87.69} \\
\bottomrule
\end{tabular}
\end{table}

We conduct an ablation study by incrementally adding each module to STGCN++ using HumanAct12 with 100\% data usage:
As shown in \Cref{tab:augment}, incorporating condition embedding and classification loss yielded an accuracy improvement of 2.35\%. Although Sampling Dropout enhanced diversity, it also introduced fidelity degradation issues when used independently. However, this issue was effectively mitigated by the Generative Refinement Module, which enabled our method to maintain diversity and accuracy, as evidenced by the highest performance of 83.19\%  achieved using all proposed components.

\begin{table}[htbp]
\centering
\small
\caption{Accuracy of Different Data Augmentation Methods on HumanAct12 for Skeleton-Based Action Recognition.}
\label{tab:diffaug}

\begin{tabular}{c|cccccc}
\toprule
\textbf{Ratio} & \textbf{100\%} & \textbf{95\%} & \textbf{90\%} & \textbf{75\%} \\
\midrule
W/O Augmentation & 75.69 & 77.78 & 75.00 & 73.61  \\
Gaussian Noise & 78.47 & 79.17 & 79.86 & 79.86 \\
Scaling  & 79.17 & 77.08 & 75.00 & 72.22  \\
Rotating    & 79.17 & 77.08 & 79.86  & 75.00 \\ \midrule
Ours & \cellcolor{green!20}\textbf{81.54}       & \cellcolor{green!20}\textbf{80.77}      &    \cellcolor{green!20}\textbf{80.00}  &   \cellcolor{green!20}\textbf{80.56} \\
\bottomrule
\end{tabular}
\end{table}
\subsubsection{Dropout and Renoise Strategy Comparison}
We evaluated the impact of different dropout rates and GRM renoise values on model performance. As shown in \Cref{tab:dropout}, we tested four dropout settings during the sampling stage: no dropout, 0.1, 0.2, and 0.5. The results indicate that higher dropout values increase diversity but also lead to a loss of fine-grained details. A dropout rate of 0.2 achieved the best performance for downstream skeleton-based action recognition with STGCN++ during sampling, suggesting that introducing a moderate dropout value can enhance the diversity of generated skeletal motions while also improving recognition accuracy.

\begin{table}[htbp]
\centering
\caption{Comparison of skeleton generation quality on the HumanAct12 dataset. }
\resizebox{\linewidth}{!}{ 
\begin{tabular}{l|c|c|c|c|c}
\toprule
\textbf{Method}         & \textbf{FID $\downarrow$} & \textbf{KID $\downarrow$} & \textbf{Diversity $\uparrow$} & \textbf{Precision$\uparrow$} & \textbf{Recall$\uparrow$}  \\ \midrule
Real data      &  0.8398   & 0.0001    &  7.217         & 0.996 & 0.999              \\ \midrule
MDM-orig \cite{tevet2023human}       &   11.3120  &  0.0512   &     6.223      &     0.996 &0.896             \\
MDM            & 24.9942     &  0.1168   &    3.7580       &    0.998& 0.390              \\
T2M-GPT\cite{zhang2023t2mgptgeneratinghumanmotion} &2.0362&\cellcolor{green!20}\textbf{0.0057}&6.6808&\cellcolor{green!20}\textbf{0.999}&0.980 \\ \midrule
\textbf{\makecell{Ours}} &   \cellcolor{green!20}\textbf{1.3288}  & 0.0170    &   \cellcolor{green!20}\textbf{6.8087} &  0.996& \cellcolor{green!20}\textbf{0.994}            \\ \bottomrule
\end{tabular}
}
\label{tab:generative}
\end{table}
As shown in Table \ref{tab:grm}, we further examined the Generative Refinement Module (GRM) under different renoise values. Here, renoise refers to a threshold such that generated samples with deviations larger than this value from the original data are discarded. On the HumanAct12 dataset, using renoise values between 10 and 20 proved effective in filtering out distorted samples, thereby reducing their negative impact on downstream action recognition tasks. Notably, in the downstream main results, we adopt unified dropout and GRM values and report the average over five repeated runs, which yields more robust results across tasks. Although both sets of parameters require manual tuning, they exhibit a certain degree of robustness: even when the chosen values are not optimal, the model still surpasses the baseline in downstream performance.

\begin{figure*}[th!]
    \centering
    \includegraphics[width=0.99\linewidth]{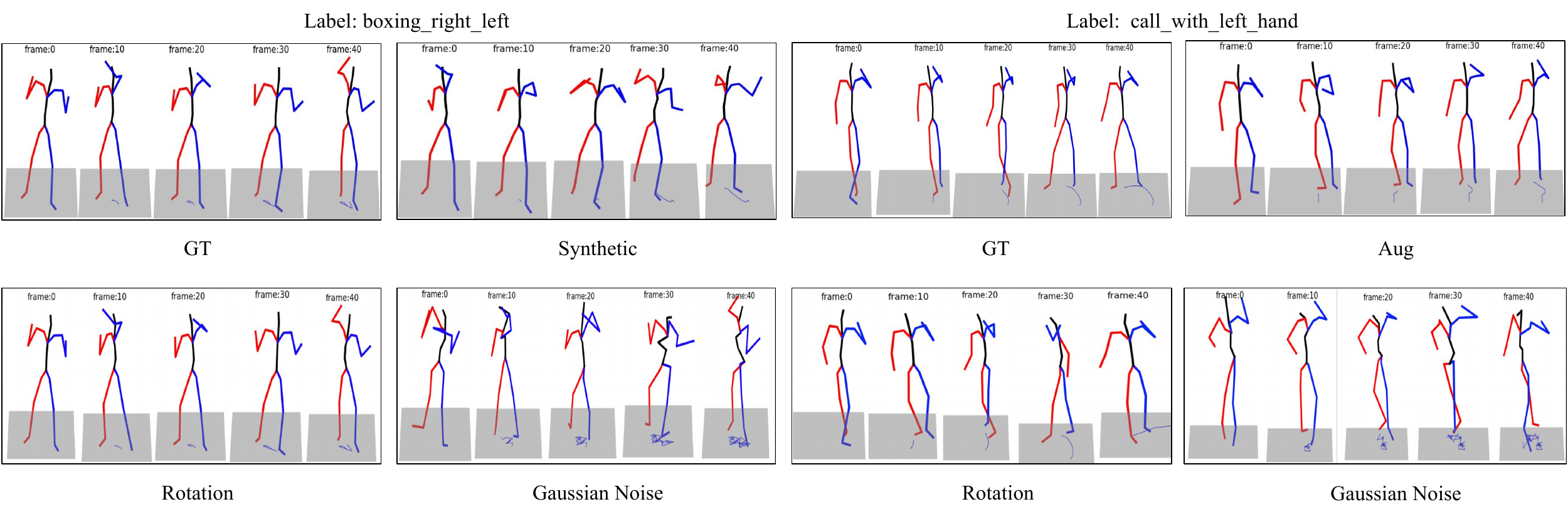}
    \caption{Visualisation of our generated skeleton sequences conditioned on action labels. Our results demonstrate that our method generates diverse motion patterns while preserving label-specific semantics. Additional visualisations are provided in the supplementary file.}
    \label{fig:vis}
\end{figure*}
\subsubsection{Comparison of Data Augmentation Methods}
We compared the effectiveness of different data augmentation methods on downstream tasks in \Cref{tab:diffaug}. Unlike conventional methods, which indiscriminately perturb motion data without considering semantic consistency, our conditional generative approach produces label-consistent and realistic motion variations, enhancing diversity and performance, particularly in low-data settings.

\subsubsection{Qualitative Results}

\Cref{fig:vis} presents visualisations of conditional skeletal generation results. By analysing these qualitative results, we highlighted the diversity and fidelity of the generated samples. The results illustrated how our method preserved label-specific semantics while introducing subtle variations in key joints relevant to action recognition, in contrast to conventional skeleton data augmentation methods that mainly rely on simple geometric transformations. Moreover, the visualisations indicate that our generated data carries clear physical meaning, such as variations in movement speed and joint angles, since the 263-dimensional inputs encode rich physical attributes rather than mere skeleton points. Additional visualisations are provided in the supplementary material.

\section{Conclusion}

We presented a conditional diffusion framework for skeleton-based action recognition, generating diverse, label-consistent motion sequences. Our approach significantly improves recognition performance, especially under limited data conditions, and consistently benefits a range of skeleton action recognition backbone architectures. We demonstrated how our design balances fidelity and diversity through extensive ablations, enabling scalable and controllable augmentation. This work represents a significant step forward in generative augmentation for structured human motion data, with practical implications for data-efficient learning in action recognition tasks and for reducing the cost of collecting large-scale skeleton data.

\paragraph{Limitations and Future Work} While our method generalises well across datasets and models, generation quality may degrade for rare or ambiguous actions under extreme label imbalance. As future work, we will focus on incorporating kinematic priors—such as joint angle constraints and temporal smoothness filters. Furthermore, we will investigate adaptive tuning mechanisms for key hyperparameters (e.g., GRM thresholds and dropout rates) and uncertainty-aware conditioning to enhance robustness, while extending the model to multi-person interactions and longer motion sequences.

{\small
\bibliographystyle{ieee}
\bibliography{egbib}
}

\end{document}